\documentclass{article}

\usepackage[nonatbib,preprint]{neurips_2025}
\usepackage[numbers]{natbib}

\usepackage[utf8]{inputenc} %
\usepackage[T1]{fontenc}    %
\usepackage{hyperref}       %
\usepackage{url}            %
\usepackage{booktabs}       %
\usepackage{amsfonts}       %
\usepackage{nicefrac}       %
\usepackage{microtype}      %
\usepackage{xcolor}         %

\usepackage{lineno}

\usepackage{graphicx}

\usepackage{subfigure}
\usepackage{siunitx}

\usepackage{amsmath}
\usepackage{amssymb}
\usepackage{mathtools}
\usepackage{amsthm}
\usepackage{float}
\usepackage{multirow}
\usepackage{tablefootnote}

\definecolor{darkblue}{rgb}{0, 0, 0.5}
\hypersetup{colorlinks=true, citecolor=darkblue, linkcolor=darkblue, urlcolor=darkblue}

\title{SWAN-GPT: An Efficient and Scalable Approach for \\ Long-Context Language Modeling}

\author{%
  Krishna C. Puvvada\thanks{Equal contribution. Correspondence emails should be sent to: \texttt{\{kpuvvada, fladhak\}}@nvidia.com} \\
  \And
  Faisal Ladhak\footnotemark[1] \\
  \And
  Santiago Akle Serrano \\
  \AND
  Cheng-Ping Hsieh \\
  \And
  Shantanu Acharya \\
  \And
  Somshubra Majumdar \\
  \And
  Fei Jia \\
  \And
  Samuel Kriman \\
  \And
  Simeng Sun \\
  \And
  Dima Rekesh \\
  \And
  Boris Ginsburg \\
  \AND
  \textbf{NVIDIA}
}

\begin{document}
\maketitle

\begin{abstract}
We present a decoder-only Transformer architecture that robustly generalizes to sequence lengths substantially longer than those seen during training. Our model, SWAN-GPT, interleaves layers without positional encodings (NoPE) and sliding-window attention layers equipped with rotary positional encodings (SWA-RoPE). Experiments demonstrate strong performance on sequence lengths significantly longer than the training length without the need for additional long-context training. This robust length extrapolation is achieved through our novel architecture, enhanced by a straightforward dynamic scaling of attention scores during inference. In addition, SWAN-GPT is more computationally efficient than standard GPT architectures, resulting in cheaper training and higher throughput.  Further, we demonstrate that existing pre-trained decoder-only models can be efficiently converted to the SWAN architecture with minimal continued training, enabling longer contexts. Overall, our work presents an effective approach for scaling language models to longer contexts in a robust and efficient manner.

\end{abstract}

\section{Introduction}
\label{introduction}

Large Language Models based on standard decoder-only transformer architectures \citep{brown2020languagemodelsfewshotlearners,grattafiori2024llama,yang2025qwen2} struggle with context lengths beyond their training distribution. Current approaches to extending context length fall into two categories: specialized training on increasingly longer sequences \citep{grattafiori2024llama, yang2025qwen2, peng2023yarn, chen2023extendingcontextwindowlarge} or complex inference time modifications \citep{an2024trainingfreelongcontextscalinglarge}. These approaches incur either increased computation cost or increased implementation complexity. We propose SWAN-GPT, a decoder-only transformer architecture that natively handles sequences substantially longer than seen during training without requiring additional long-context-specific training. By strategically interleaving global attention layers without positional encodings and local, sliding-window attention layers with rotary position encodings, combined with a dynamic attention scaling mechanism, SWAN-GPT achieves both remarkable length generalization and significant computational efficiency. This architecture not only maintains comparable performance to standard transformers on established LLM benchmarks, but also achieves robust extrapolation to sequences far beyond the training length, providing a more scalable and efficient solution to the long-context challenge.

A central challenge in extending transformer context lengths is the handling of positional information. Transformers rely on positional encodings to track token order, but these encodings often become unreliable when models process sequences longer than those seen during training. Among the various positional encoding schemes, Rotary Positional Encodings (RoPE) \citep{RoFormer} have been widely adopted in modern language models due to effectiveness in capturing relative positions. However, RoPE-based models exhibit significant performance degradation when applied to sequences exceeding their training length. This degradation occurs because inter-token distances advance to ranges where the relative rotation angle is outside the trained distribution \citep{liu2024}. 

To address this limitation, we explore two complementary approaches with distinct strengths and limitations. Sliding window attention with RoPE (SWA-RoPE) restricts every token's attention to a fixed-size window of neighboring tokens. Because the distance between attended tokens is bounded, SWA-RoPE layers never operate at rotation angles outside their training range, making them inherently robust to arbitrary sequence lengths. However, this locality constraint limits their ability to capture long-range dependencies. Conversely, layers without positional encoding (NoPE) \citep{haviv2022,kazemnejad2023impact} allow unrestricted attention across the entire context while omitting explicit positional information. Notably, autoregressive NoPE models can develop implicit positional awareness through the causal attention mask, achieving comparable perplexity to models with explicit positional embeddings \citep{haviv2022}. Despite this capability, pure NoPE models also exhibit poor robustness beyond their training length, with performance degrading rapidly due to the brittleness of the learned positional mechanism.

Our key insight is that these approaches can complement each other through strategic integration. SWAN-GPT interleaves global attention layers without positional encodings (NoPE) and local sliding-window attention layers with rotary positional encodings (SWA-RoPE). This hybrid design creates a synergistic effect: SWA-RoPE layers provide local positional structure, while NoPE layers integrate information across arbitrary distances. When interleaved, the NoPE layers develop more robust representations than they would in isolation, enabling the entire model to generalize beyond its training sequence length. Unlike standard RoPE-based transformers which experience catastrophic performance collapse outside their training context, SWAN maintains robust performance on extended sequences with only a straightforward rescaling of attention scores during inference.

In Section 2.1, we provide evidence that failures in the implicit position prediction mechanism of NoPE models contribute to their performance degradation on longer sequences, and demonstrate how the interleaved SWA-RoPE layers stabilize this mechanism. Additionally, we show that existing transformer models can be efficiently adapted to the SWAN architecture through continued pretraining (CPT), offering a practical, cost-effective path to upgrading deployed models.

Our contributions are as follows:
\begin{enumerate}
    \item A novel architecture (SWAN) that combines SWA-RoPE and NoPE layers to enable efficient length extrapolation without additional training, enhanced by a logarithmic attention scaling mechanism for inference.
    \item Mechanistic analysis explaining why this architecture produces robust length extrapolation, with evidence that NoPE layers develop more stable positional encodings when paired with SWA-RoPE layers.
    \item Empirical results demonstrating that SWAN maintains robust performance on sequences far exceeding its training length, while achieving comparable results to standard transformer architectures on established LLM benchmarks.
    \item A practical method for adapting existing transformer models to the SWAN architecture through continued pre-training (CPT), providing a cost-effective upgrade path for deployed models.
\end{enumerate}

\section{The SWAN-GPT architecture}
SWAN-GPT is a decoder-only Transformer architecture that addresses the challenge of length extrapolation by interleaving two types of attention mechanisms: global attention layers without positional encodings (NoPE) and local sliding-window attention layers with rotary positional encodings (SWA-RoPE). This hybrid design leverages the complementary strengths of both approaches to achieve robust length extrapolation capabilities,  without specialized long-context training. 

As detailed in our ablation studies (\autoref{sec:ablations}), we explored multiple configurations for interleaving these layer types. Our experiments revealed that beginning with a global NoPE layer followed by three consecutive sliding-window layers, repeating this pattern throughout the network, demonstrated superior performance on long-context tasks. This configuration achieves exceptional NIAH scores at context lengths $16$ times longer than the training length, and maintains robust performance even at $32$ times the training length when combined with appropriate attention scaling (\autoref{sec:attn_scaling}). We adopt this interleaving pattern for all experiments presented in the main paper.

The global NoPE layers permit unrestricted attention across the entire context, enabling the model to capture long-range dependencies. Meanwhile, the local SWA-RoPE layers operate with a fixed window size of $512$ tokens, providing consistent positional information within a bounded context. This architecture creates a complementary system where SWA-RoPE layers enforce local positional structure while NoPE layers integrate information across arbitrary distances. The key insight is that when these mechanisms are interleaved, the NoPE layers develop more robust position-aware representations than they would in isolation, enabling the entire model to generalize effectively beyond training sequence lengths.

\begin{figure}
    \centering
    \includegraphics[width=0.7\linewidth]{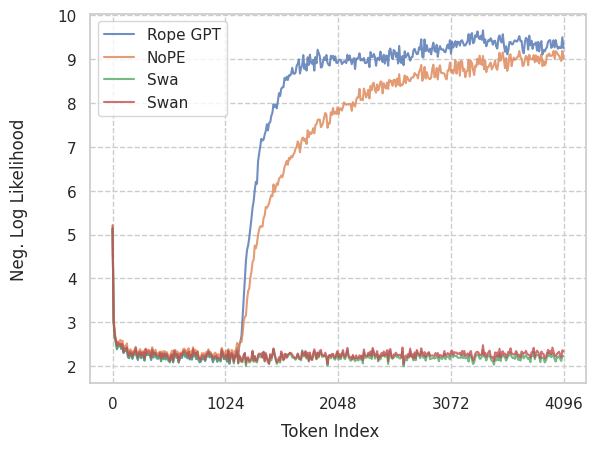}
    \caption{Mean negative log likelihood by token position for GPT with rotary positional encodings (RoPE GPT, blue), a GPT with no positional encodings (NoPE, orange), a Swan model (red), and a model conposed only of sliding window attention layers (SWA, green). Both RoPE GPT and NoPE models struggle beyond training sequence length (1024). SWA model doesn't experience such catastrophic failure due to its limited context. Swan model behaves like a SWA model without the limitation of SWA model due to its global NoPE layers.}
    \label{fig:mean_rank}
\end{figure}

Figure \ref{fig:mean_rank} demonstrates this capability by comparing four models trained on sequences of up to $1024$ tokens: a standard GPT model with RoPE, one with no positional encodings (NoPE), one with only sliding window attention (SWA) and one using our architecture (SWAN). We evaluate the model's predictions on $1280$ validation sequences of length $4096$. The plot shows the negative log likelihood at each sequence position averaged over all validation sequences, with lower values indicating better performance. Both the RoPE and NoPE models experience significant performance degradation beyond their training length , with negative log likelihood increasing sharply beyond 1024 tokens. In contrast, both the SWAN and SWA architectures maintain consistent predictive quality throughout the entire 4096-token range, demonstrating their robust length extrapolation capabilities. Notably, SWAN maintains this performance while retaining the ability to capture long-range dependencies that the purely local SWA approach cannot (see \autoref{sec:ablations}).

In the following sections, we examine why this architecture works so effectively for length extrapolation, providing mechanistic analysis and empirical evidence of its robust performance.

\subsection{Stabilizing Implicit Position Encodings for Robust Length Extrapolation}
\label{section:mechanistic}
A key question in our investigation is understanding why the NoPE layers within our SWAN architecture demonstrate substantially more robust length extrapolation capabilities compared to identical layers within a model built purely of NoPE layers.

Despite the absence of explicit positional encoding, prior work has demonstrated that trained NoPE models implicitly learn to predict token positions after processing through a few layers \citep{Chi2023}. 
This implicit position embedding emerges from the autoregressive nature of decoder-only models, where tokens later in the sequence have access to more context than earlier tokens, creating distinct distributions at different positions. 
These distributional differences enable NoPE models to infer positional information and incorporate it into their predictions \citep{Chi2023}.

However, standard NoPE models exhibit poor robustness to sequences longer than their training length, with performance degrading rapidly beyond the training boundary. 
We hypothesize that this limitation stems from a failure in their implicit position prediction mechanism when extrapolating to longer contexts. We further hypothesize that in our SWAN architecture, the interleaved SWA-RoPE layers relieve the NoPE layers from needing to develop the brittle position encodings seen in pure NoPE models, resulting in more robust processing of longer sequences.

To test these hypotheses, we conducted experiments with both pure NoPE and SWAN models trained on sequences of $1024$ tokens and evaluate them on sequences of $2048$ tokens. We employed two complementary analysis techniques: (1) position prediction probes to quantitatively measure positional information in model representations, and (2) attention pattern visualization to examine how attention mechanisms behave when processing sequences beyond training length.

\subsubsection{Position Prediction Probes}
\label{subsub:probes}
\begin{figure}
    \centering
    \includegraphics[width=0.75\linewidth]{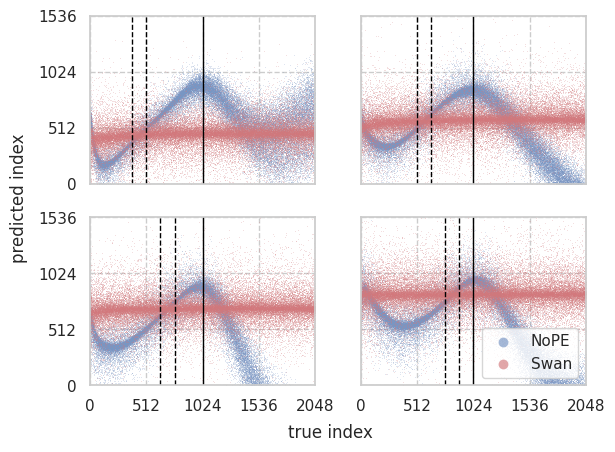}
    \caption{Predictions of token indices by 8 different probes. Each probe is trained with tokens from one model and different context regions (demarcated by dashed lines). Probes on NoPE models (blue) extrapolate correctly up until the maximum NoPE training length (solid line). Probes on SWAN (red) are not predictive of token indices.}
    \label{fig:probe_invalidation}
\end{figure}

To provide evidence for our hypothesis, we trained probes that predict token positions from token embeddings. We evaluated these probes on held-out tokens from positions both within and beyond the models' training range. Figure \ref{fig:probe_invalidation} shows predictions from eight different probes, each trained with tokens sampled from ranges demarcated by dashed lines. Each of the four subplots shows results from two probes - one trained on NoPE model embeddings (blue) and one on SWAN model embeddings (red) - with each probe trained on tokens from different context regions demarcated by dashed lines.

For pure NoPE models (blue points), the probe's predictions extrapolate well up to the boundary of the model's training range (solid black line). However, probes cease to be predictive beyond the training boundary. Furthermore, probes trained in different sub-regions of the range all fail at the same location. This phenomenon is consistent with the notion that the position prediction mechanism in NoPE models fails beyond the model's training range. In contrast, the SWAN model (red points) exhibits fundamentally different behavior. Position probes trained on SWAN's NoPE layers show little positional information across all sequence positions, suggesting these layers do not develop the same brittle position encoding mechanism seen in pure NoPE models. This supports our hypothesis that the interleaved SWA-RoPE layers stabilize the NoPE layers by relieving them from the need to track absolute positions, instead allowing them to focus on integrating information across arbitrary distances while the SWA-RoPE layers handle local positional structure.

\subsubsection{Attention Pattern Analysis}

To further investigate this phenomenon we examine the average attention values at different token positions for different sequence lengths. We average the probability scores (attention scores post soft-max) over all heads and over a set of validation batches. We randomize the token order in order to remove the effect of the correlation structure present in natural language. 

\begin{figure}[tp]
    \centering
    \includegraphics[width=0.8\linewidth]{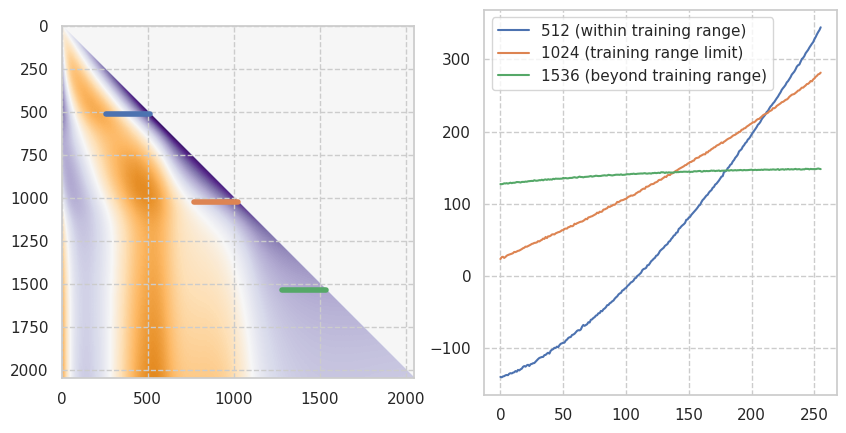}
    \caption{Attention maps for 6th layer of NoPE model. Averaged over all heads and all validation records (left). Cross section for sequence of length 512, 1024 (limit of model training range) and 1536 in length extrapolation regime. Attention pattern of leading 256 tokens differ for sequences within and beyond training range.}
    \label{fig:nopos-avg-attn}
\end{figure}

\autoref{fig:nopos-avg-attn} displays the average attention maps of the 6th layer in the NoPE model. The visualization reveals that for sequences longer than the training length (green) the model places roughly the same amount of attention to all of the 256 tokens preceding the target token. Whereas for sequences within the training range (orange and blue) it preferentially attends to the tokens closest to the target token. A model that extrapolates to longer sequences should maintain a similar attention pattern for tokens close to the target token, regardless of sequence length. 

\begin{figure}[ht]
    \centering
    \includegraphics[width=0.8\linewidth]{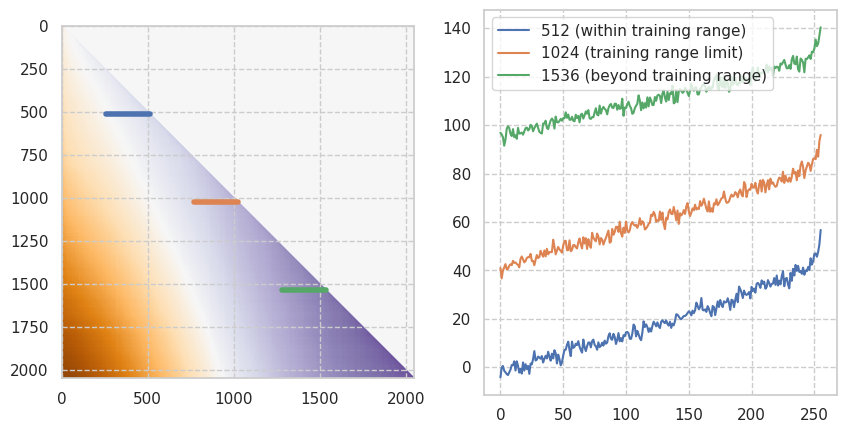}
    \caption{Attention maps for 20th layer of our SWAN model (6th NoPE layer). Averaged over all heads and all validation records (left). Cross section for sequence of length 512, 1024 (limit of model training range) and 1536 in length extrapolation regime. Attention pattern of leading 256 tokens show consistent decay patterns for sequences with length within and beyond training range.}
    \label{fig:swan-avg-attn}
\end{figure}

In contrast, \autoref{fig:swan-avg-attn} displays the average attention maps of the 20th layer (the 6th NoPE layer) of our SWAN model. Unlike the pure NoPE model, SWAN's attention maps exhibit consistent attention patterns for sequences with lengths within and beyond the training regime.

These analyses support our hypothesis that interleaving SWA-RoPE layers fundamentally alters how NoPE layers process positional information. The use of positional embeddings in the SWA-RoPE layers appears to stabilize the representations in the NoPE layers, making them more robust to sequence length extrapolation. This suggests that SWAN's superior length extrapolation capability stems from the emergent properties of the interleaved architecture.

\subsection{Dynamic Attention Scaling for Extended Context Processing}
\label{sec:attn_scaling}

\begin{figure*}[bp]
    \centering
    \includegraphics[width=0.9\textwidth]{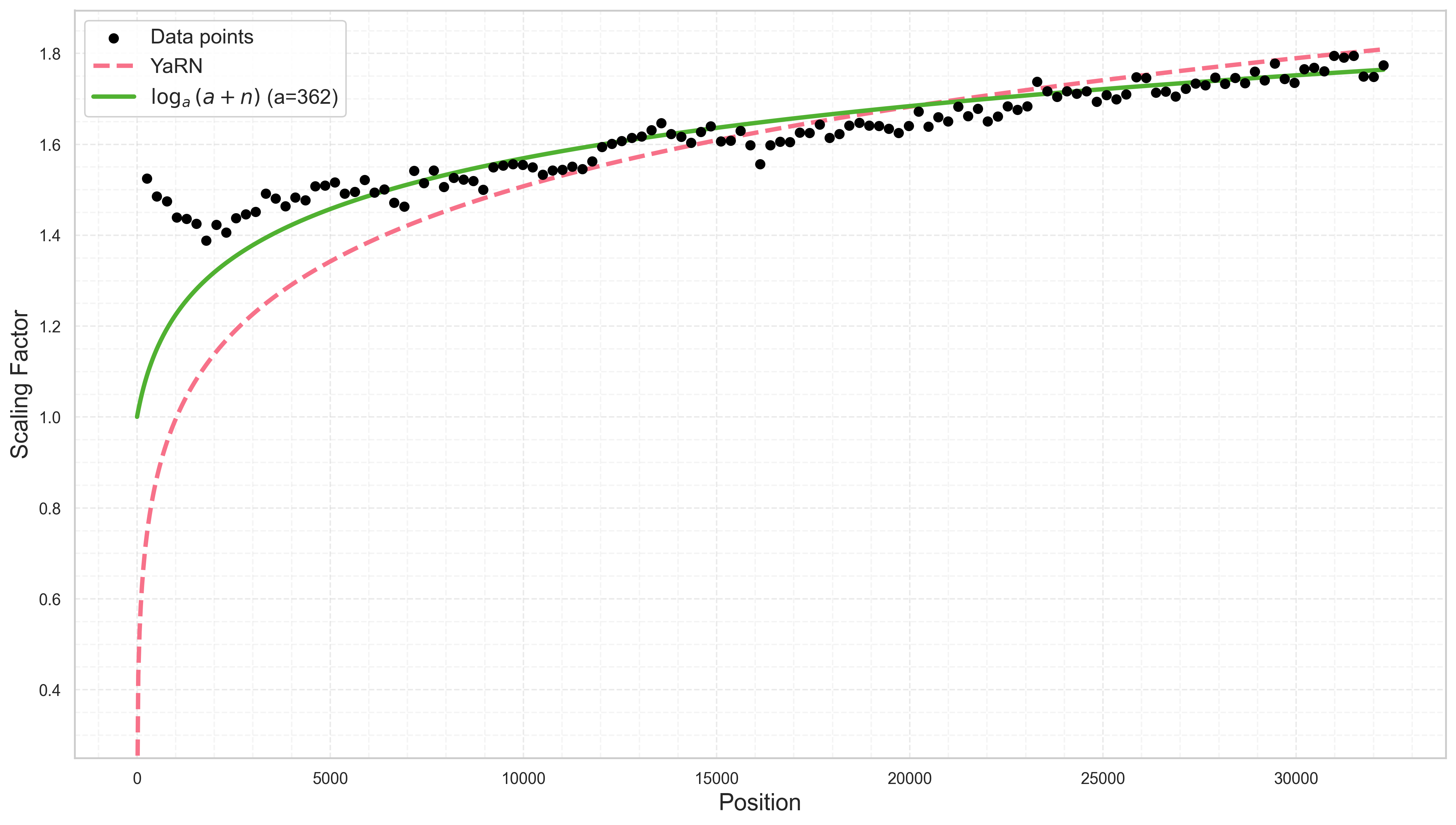}
    \caption{Estimates of optimal scaling factors (black) comparing the fit of our logarithmic scaling function vs. YaRN scaling. We find that YaRN scaling doesn't work as well for NoPE layers.}
    \label{fig:scaling}
\end{figure*}

While our architecture demonstrates inherent sequence length extrapolation, we find that further performance improvements can be achieved through proper scaling of attention logits during inference. This scaling is particularly important for the global NoPE layers, which must effectively integrate information across arbitrary distances as sequence length increases.

Prior work has shown that RoPE-based models improve their performance on extended context lengths when the temperature of the attention logits is properly adjusted \citep{peng2023yarn}. The SWA-RoPE layers in our SWAN architecture inherently handle longer sequences due to their local attention window. However, we hypothesize that the global attention NoPE layers may still require scaling to maintain performance at extended lengths. 

For this analysis, we sampled 200 documents from the model's training distribution (each with at least 32K tokens) to maintain consistent semantic distribution while extending context length beyond the original 1K tokens used during training. We partitioned each 32K-token context into 128-token windows and estimated a single optimal scaling factor per window by minimizing its perplexity over all 200 documents.

\autoref{fig:scaling} shows the empirically determined optimal scaling factors (black dots) across different positions in the 32K context. We find that a logarithmic scaling function $\log_a(a+n)$ (green line) provides an excellent fit to the empirical data. This function captures two key properties we observe -- a natural growth rate that matches the data's progression, and a base scaling factor that never falls below 1.0, which is important for maintaining model stability at early positions. Interestingly, while prior work found that the YaRN scaling function \citep{peng2023yarn} works well for RoPE-based models, we observe that it (dashed pink line) fits poorly for the NoPE layers in our SWAN architecture, particularly in early positions where it significantly under-estimates the required scaling.

\begin{figure*}[t]
    \centering
    \includegraphics[width=0.9\textwidth]{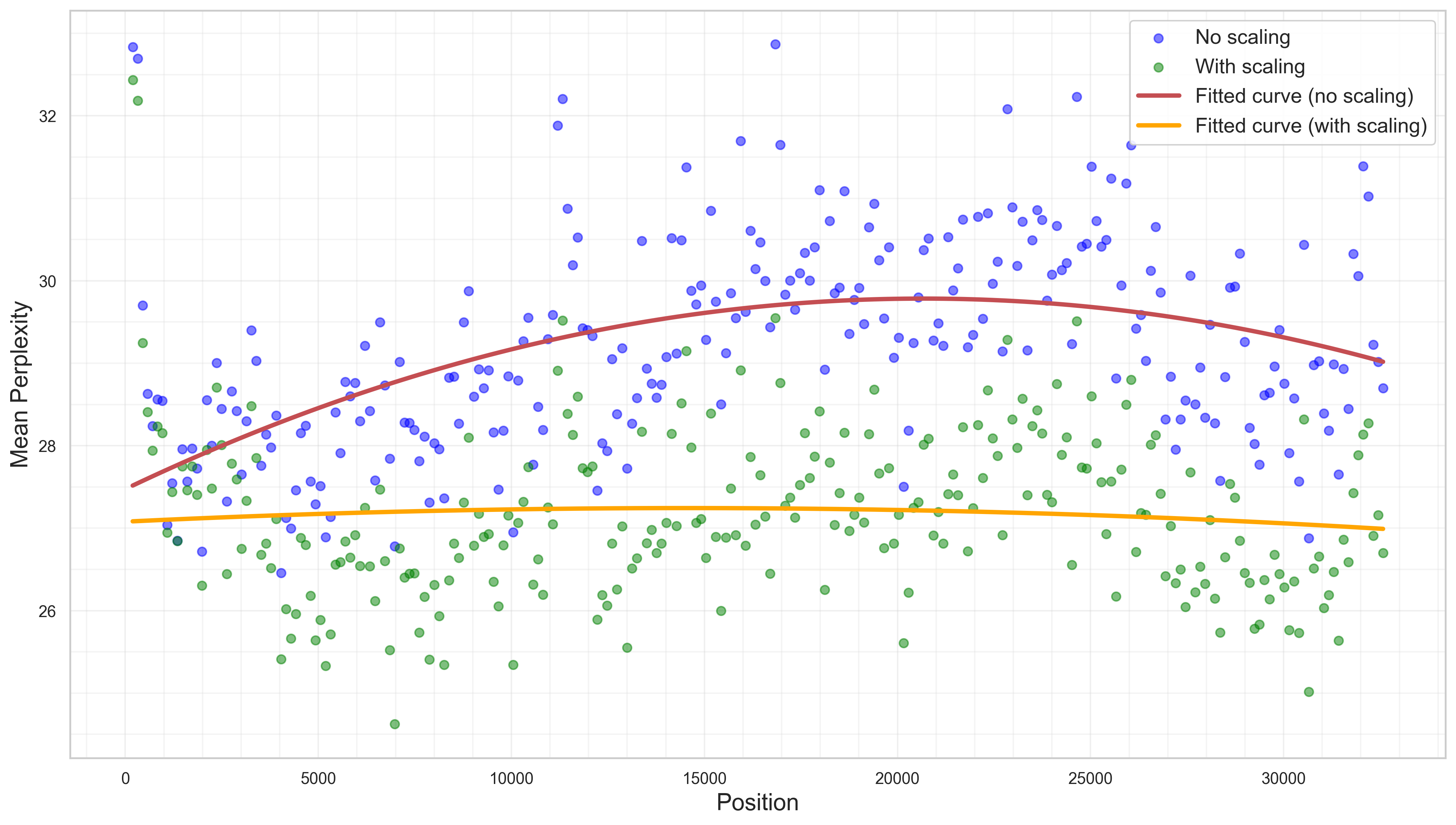}
    \caption{Perplexity on held-out documents, with (green) and without (blue) logarithmic scaling applied to attention scores.}
    \label{fig:ppl_pg19}
\end{figure*}

Having identified a suitable logarithmic scaling function, we next investigate whether applying this scaling would improve model performance on unseen data. To validate our approach we use held-out documents from the PG19 dataset. We compute the average perplexity for each $128$ token window in documents with $32$K tokens. \autoref{fig:ppl_pg19} plots the perplexity at each location within the 32K token context, with and without our scaling function applied. Without scaling (blue), we observe a clear degradation in model performance on longer contexts. In contrast, our scaling (green points) allows the model to maintain better performance as measured by a lower and more stable perplexity value for the entire context length up to contexts $32$ times longer than the training length ($1K$ tokens). This improved performance with scaling is further validated by our NIAH evaluation results in \autoref{tab:model-comparison} in \autoref{sec:ablations}, where we demonstrate that scaling improves NIAH scores from $0.171$ to $0.957$ at 8K context length and from $0.005$ to $0.907$ at 16K context length.

\section{Results}

In the previous section, we introduced the SWAN architecture and motivated its robust length extrapolation via mechanistic analysis and empirical experiments. Here, we evaluate the effectiveness of the proposed architecture compared to standard RoPE-based transformer LLMs. Our goal is to demonstrate that SWAN models can maintain similar performance on standard LLM benchmarks while achieving substantially improved length extrapolation capabilities beyond the training context length.

We trained both RoPE GPT and SWAN models with 1B parameters from scratch using 1T tokens at 8K sequence length with a token batch size of 6M. The SWAN model followed 1:3 global:local ratio, with sliding window attention layers using a 512-token window size. We evaluated both models on standard LLM benchmarks using the LM Evaluation Harness Library \citep{eval-harness}. As shown in \autoref{tab:standardbench}, the SWAN model performs comparably or better than the RoPE model across all benchmarks, achieving an average $51.4\%$ vs. $49.5\%$. 

\begin{table*}[t]
\centering
\begin{tabular}{l|rrrrrrrr|r}
\toprule
\textbf{Dataset} & \textbf{ARC-E} & \textbf{ARC-C} & \textbf{H} & \textbf{W} & \textbf{RACE} & \textbf{PIQA} & \textbf{SIQA} & \textbf{OBQA} & \textbf{Avg} \\
\midrule
RoPE & 65.36 & 38.23 & 58.35 & 57.93 & 35.02 & 73.12 & 32.91 & 35.20 & 49.5 \\
SWAN & 69.40 & 41.04 & 59.76 & 59.75 & 35.69 & 73.99 & 33.73 & 37.80 & 51.4 \\
\bottomrule
\end{tabular}
\caption{Results for 1B models trained on 1T tokens. The models were evaluated on ARC-Easy, ARC-Challenge, Hellaswag, Winogrande, RACE, PIQA, Social IQA, and Openbook QA. The SWAN model shows comparable or better performance across all benchmarks.}
\label{tab:standardbench}
\end{table*}

The primary advantage of the SWAN architecture becomes evident when evaluating its performance on sequences significantly longer than those seen during training. \autoref{tab:ruler_1b} shows the results for both models on the Ruler benchmark \citep{hsieh2024ruler} across various context lengths. While both models get similar performance for sequence lengths within their training distribution ($\leq8K$), their behaviors diverge dramatically beyond this point. The standard RoPE based model fails completely when presented with sequences exceeding its training length, showing catastrophic degradation. In contrast, the SWAN model exhibits a much more graceful degradation pattern even at sequences substantially longer than the training length.

\begin{table*}[tp]
\centering
\begin{tabular}{l|c|ccccccc}
\toprule
\textbf{Model} & \textbf{MTL} & \textbf{4K} & \textbf{8K} & \textbf{16K} & \textbf{32K} & \textbf{64K} & \textbf{128K} & \textbf{256k} \\
\midrule
RoPE GPT-1B   & 8K & 70.6 & 53.5 & NA & NA & NA & NA & NA \\
Swan GPT-1B   & 8K & 68.1 & 52.4 & 45.8 & 36.9 & 30.6 & 24.4 & 14.9\\
\bottomrule
\end{tabular}
\caption{Comparing long-context performance of SWAN-GPT with standard RoPE-based models on the Ruler benchmark. MTL=Maximum training length. The SWAN model maintains measurable performance even at 32× its training length, while the RoPE model fails completely beyond its training length.}
\label{tab:ruler_1b}
\end{table*}

\subsection{Efficient Adaptation of Pre-trained Models to SWAN Architecture}
\label{sub:pre-trained_swan_adaptation}

While training models from scratch demonstrates that our architecture achieves comparable results to RoPE-based transformers on standard benchmarks while offering superior length extrapolation, adapting existing pre-trained models would significantly enhance the practical utility of our approach.

Prior research has established that most of the knowledge in transformer models is encoded in the feed-forward layers, with attention mechanisms primarily serving to route information \citep{geva-etal-2021-transformer}. 
Since SWAN primarily modifies the attention computation while preserving feed-forward layers, we hypothesize that existing pre-trained models can be efficiently converted to the SWAN architecture without losing their accumulated knowledge.
This adaptation capability would make our approach immediately applicable to the large ecosystem of existing transformer models, offering a cost-effective path to enhanced length extrapolation without full retraining.

We start with an 8B parameter RoPE GPT model that was pre-trained for 15T tokens context length of 8K tokens \citep{su2024nemotroncctransformingcommoncrawl}. We converted this model to the SWAN architecture by initializing all weights from the pre-trained RoPE GPT model and modifying the attention layers to implement our 1:3 global-local pattern as established in Section 2. This process involved removing positional encodings from global attention layers, configuring sliding-window attention with a window size of 512 tokens in local layers. Following initialization, we performed continued pre-training (CPT) for an additional 315B tokens (approximately $2\%$ of the original pre-training compute) at an extended context length of 32K tokens. The process utilized the same data distribution as the original model, with sequence lengths extended to 32K through concatenation of shorter examples. For the final 15B tokens, we applied Fill-in-Middle augmentation \citep{bavarian2022efficienttraininglanguagemodels} to further enhance the model's contextual understanding.

Post-training for RoPE GPT model was conducted in two stages, with the first stage focusing on math and code followed by a general SFT in the second stage. Post-training for Swan followed similar procedure, but with the sequence length extended to 32K through concatenation of shorter examples. To enhance long-context capabilities, we augmented the SFT training data with a variety of tasks designed to exercise the model's ability to reason over extended contexts. These included questions referring to previous turns in concatenated examples and synthetic tasks such as filling in the middle, recalling portions of context based on keywords, tracing linked lists, executing basic SQL queries on made-up table data, and multi-hop reasoning \citep{chen2024essential} tasks modified to 32K sequence length.

\begin{table*}[tp]
\centering
\begin{tabular}{ll|r|r}
\toprule
\textbf{Category} & \textbf{Benchmark} & \textbf{RoPE GPT} & \textbf{SWAN GPT} \\
\midrule

\multirow{2}{*}{Math}
 & GSM8k                & 87.7  & 87.7 \\
 & MATH500              & 70.4  & 68.4 \\
\midrule

\multirow{4}{*}{Code}
 & MBPP                 & 76.2  & 75.7 \\
 & MBPP+                & 66.1  & 65.3 \\
 & HumanEval            & 74.4  & 75.0 \\
 & HumanEval+           & 68.3  & 68.3 \\
\midrule

\multirow{4}{*}{General}
 & MT-Bench             & 7.35  & 7.43 \\
 & MMLU (generative)    & 68.0  & 65.4 \\
 & IFEval (Prompt)      & 63.0  & 62.7 \\
 & IFEval (Instruction) & 72.7  & 72.2 \\
\midrule

\multirow{1}{*}{Tool Use}
 & BFCL v2 Live         & 68.7  & 68.9 \\
\midrule

\multirow{1}{*}{Long Context}
 & RULER (128k context)         & NA  & 77.8 \\
\midrule

\multicolumn{2}{r|}{\textbf{Average (w/o MT-Bench, RULER)}} & \textbf{71.55} & \textbf{70.95} \\
\bottomrule
\end{tabular}
\caption{Comparison of RoPE GPT vs. Swan when adapting a pre-trained RoPE GPT model to SWAN model. SWAN maintains comparable performance on short benchmarks (on average) while attaining long-context capabilities.}
\label{tab:shortbench}
\end{table*}

\autoref{tab:shortbench} compares our adapted SWAN GPT-8B model with the original RoPE GPT-8B model across standard LLM benchmarks. The results demonstrate that the SWAN adaptation maintains comparable performance across a diverse set of tasks, including mathematical reasoning (GSM8k, MATH500), coding (MBPP, HumanEval), and general language understanding (MMLU, IFEval, MT-Bench). Remarkably, we observe only a minimal decrease in average performance, from $71.55\%$ to $70.95\%$, confirming our hypothesis that substantial architectural modifications to the attention mechanism can be implemented with only a brief adaptation phase while preserving the model's fundamental capabilities.

\begin{table*}[tp]
\centering
\begin{tabular}{l|c|ccccccc}
\toprule
\textbf{Model} & \textbf{MTL} & \textbf{4K} & \textbf{8K} & \textbf{16K} & \textbf{32K} & \textbf{64K} & \textbf{128K} & \textbf{256k} \\
\midrule
Llama3.1-8B   & 128K & 95.5 & 93.8 & 91.6 & 87.4 & 84.7 & 77.0 & NA \\
Qwen2.5-7B-Instruct-1M & 256K & 96.8 &  95.3 & 93.0 & 91.1 & 90.4 & 84.4 & 75.3 \\
Qwen2.5-7B-Instruct & 32K & 96.7 & 95.1 & 93.7 &  89.4 & 82.3 & 55.1 & NA \\
SwanGPT-8B      & 32K & 93.8 & 90.8 & 88.1 & 84.4 & 80.5 & 77.8 & 73.2\\
\bottomrule
\end{tabular}
\caption{Comparing Long-context performance of SWAN with other models. MTL=Maximum training length. RoPE based models degrade fast with increase in sequence length where as SWAN exhibits much more graceful dropoff.}
\label{tab:ruler}
\end{table*}

The primary advantage of converting to the SWAN architecture is the substantial improvement in length extrapolation capabilities. In \autoref{tab:ruler}, we compare our adapted SWAN GPT-8B model against state-of-the-art models of similar size on the RULER benchmark \citep{hsieh2024ruler} across various context lengths. Despite being trained with a maximum context length of only 32K, our SWAN GPT-8B model demonstrates remarkable length extrapolation capabilities. At 64K tokens (2× the training length), it achieves a RULER score of 80.5; at 128K tokens (4× the training length), it maintains a score of 77.8, and even at 256K tokens (8× the training length), it achieves a respectable score of 73.2.

This robust extrapolation capability is particularly notable compared to the performance dropoff patterns observed in other models. For example, the Qwen2.5-7B-Instruct (128K) model, which was also trained with a maximum context length of 32K, shows a large drop from 82.3 at 64K tokens to 55.1 at 128K tokens. In contrast, SWAN model exhibits a much more gradual degradation, maintaining 77.8 at 128K sequence length. Even when compared to models specifically trained on longer contexts, such as Llama3.1-8B (trained up to 128K) and Qwen2.5-7B-Instruct (1M) (trained up to 256K), SWAN remains competitive. The SWAN model's score of 77.8 at 128K tokens is comparable to Llama3.1-8B's 77.0, despite Llama3.1-8B being explicitly trained at this context length and our model being trained on contexts only one-fourth as long. Similarly, our SWAN model achieves a comparable RULER score to Qwen2.5-7B-Instruc (1M) at 256K context length, despite the latter being explicitly trained on sequences eight times longer than our maximum training length.
 
These results demonstrate that the SWAN architecture enables efficient adaptation of existing pre-trained models to handle significantly longer contexts than their original training length, without sacrificing their performance on standard benchmarks. This provides a practical, compute-efficient path for upgrading deployed models to handle longer contexts without the need for full retraining.

\section{Related Work}
Extending the context length of LLMs to hundreds of thousands or millions of tokens poses significant challenges across multiple dimensions. Architecturally, standard Transformer models face limitations due to positional encoding schemes like Rotary Positional Embeddings (RoPE) that break down beyond their training distribution \citep{RoFormer, liu2024} and from the quadratic computational and memory complexity of self-attention mechanisms, particularly as Key-Value (KV) cache sizes grow with increasing sequence length \citep{kwon2023efficient,fu2024challenges, liu2025spakelongcontextlargelanguage}. From an infrastructure perspective, longer contexts strain GPU memory capacity and bandwidth, often reducing throughput \citep{patel2024nvidia, gholami2024aimemorywall}.  Furthermore, acquiring high-quality long-context training data remains challenging \citep{lv2024longwanjuan,gao2025howto}, and evaluating performance on extended contexts requires more robust benchmarks \citep{hsieh2024ruler, liu2025spakelongcontextlargelanguage}. Our work, SWAN-GPT, primarily addresses the architectural challenges by introducing an innovative architecture that enables inherent length extrapolation and computational efficiency.

Several approaches aim to extend context length purely at inference time, avoiding costly retraining or finetuning. One line of work focuses on adapting positional encodings. For RoPE-based models, techniques like NTK-aware scaling \citep{bloc97_2023_ntk, bloc97_2023_dynamic} adjust the RoPE base frequency, while Positional Interpolation (PI) \citep{chen2023extendingcontextwindowlarge} linearly downscales position indices. However, these methods can degrade performance or require careful parameter tuning \cite{an2024trainingfreelongcontextscalinglarge}. More recent training-free methods directly modify the attention mechanism. ReRoPE \citep{rerope2023} constrains relative positions, SelfExtend \citep{jin2024llmmaybe} maps unseen large relative positions to seen ones using a floor operation combined with a local attention window, and Dual Chunk Attention (DCA) \cite{an2024trainingfreelongcontextscalinglarge} decomposes attention into intra-chunk and inter-chunk components. Another line of work leverages attention sinks or windowing, such as StreamingLLM \cite{xiao2024efficient} and LM-Infinite \citep{han2024lminfinite}, which retain initial and recent tokens based on the observation that these receive high attention. Models without explicit positional encodings (NoPE) learn implicit positional information \citep{haviv2022} but also exhibit poor extrapolation beyond their training length \citep{kazemnejad2023impact, wang2024lengthgeneralization}. 

A common strategy involves adapting pre-trained models or modifying the training process. Techniques like PI \citep{chen2023extendingcontextwindowlarge} and YaRN \cite{peng2023yarn} rescale RoPE embeddings but often achieve optimal performance only after continued pre-training (CPT) or finetuning on longer sequences. CPT on progressively longer sequences \cite{xiong2023effective} is effective but computationally prohibitive for very large models or extremely long contexts. To mitigate this cost, efficient finetuning methods like LongLoRA \cite{chen2024longlora} apply parameter-efficient tuning techniques specifically for long-context adaptation.
Recent state-of-the-art LLMs often achieve long-context capabilities through pre-training and post-training that explicitly incorporates varied sequence lengths and curated long-context data. Models like the Llama 3 series \cite{grattafiori2024llama, meta2024llama3, meta2024llama3.1} and the Qwen2.5 model \cite{yang2025qwen2}, exemplify this approach, leveraging vast computational resources and sophisticated data strategies to directly train for long-context understanding.

Beyond positional encoding limitations, the quadratic complexity of self-attention and the associated KV cache size pose major efficiency bottlenecks for long contexts \cite{liu2025spakelongcontextlargelanguage}. Architectural innovations aim to reduce this complexity. Sparse attention mechanisms, used in models like Longformer \cite{beltagy2020longformerlongdocumenttransformer} and BigBird \cite{zaheer2021bigbirdtransformerslonger}, limit attention to predefined patterns (e.g., local windows plus some global tokens). Alternative architectures like State Space Models (SSMs), such as Mamba \cite{gu2024mambalineartimesequencemodeling}, and linear RNN variants like RWKV \cite{peng2023rwkv}, achieve linear or near-linear complexity in sequence length but require training from scratch. SWAN-GPT improves efficiency partly through its architecture: the SWA-RoPE layers employ local attention, which is inherently more efficient than global attention. While the NoPE layers perform global attention, techniques like Multi-head Latent Attention (MLA) \cite{deepseekai2024deepseekv2} can be applied orthogonally to reduce the KV cache size for these layers. Additionally, the various KV cache optimization techniques -- including token dropping/eviction \cite{zhang2023h2o, xiao2024efficient}, merging (e.g., via activation beacons \cite{zhang2024activation}), compression, and quantization \cite{hooper2024kvquant, liu2025spakelongcontextlargelanguage} -- can complement architectural approaches like SWAN-GPT. 

Notably, the Gemma family of models \citep{gemma2024, gemma2025} also utilizes a hybrid of sliding window and global attention, but retains RoPE positional embeddings in all layers. This contrasts with SWAN-GPT's architecture, where global layers deliberately omit explicit positional encodings (NoPE), a distinction intended to achieve superior length extrapolation. A Concurrent work \cite{yang2025ropenopeagainnew} also explores interleaving SWA-RoPE and global NoPE layers, similar to SWAN-GPT's structure, but lacks the dynamic attention scaling mechanism proposed in SWAN-GPT, which is a crucial element to maintain the performance of global NoPE layers at extended lengths. Furthermore, our work provides mechanistic analyses revealing what causes length extrapolation issues in NoPE layers and how interleaving SWA-RoPE layers addresses this problem. We further demonstrate that existing pre-trained models can be efficiently converted to SWAN architecture with minimal CPT.

\section{Conclusion}

We introduced SWAN-GPT, a decoder-only transformer architecture that achieves robust length extrapolation without specialized long-context training. By interleaving NoPE and SWA-RoPE layers, along with a dynamic attention scaling, our approach maintains consistent performance on sequences substantially longer than those seen during training. Our mechanistic analysis revealed that this hybrid architecture creates a synergistic effect, where SWA-RoPE layers provide stable positional grounding that relieves NoPE layers from developing brittle positional representations. Beyond architecture innovation, we demonstrated that existing pre-trained models can be efficiently adapted to the SWAN architecture through continued pre-training, requiring only about $2\%$ of the original training compute. This offers a practical, cost-effective path for upgrading deployed models to handle significantly longer contexts without full retraining or sacrificing performance on standard benchmarks. This approach represents a significant shift away from the current paradigm of training models directly on increasingly longer sequences, offering a more computationally efficient path toward long-context language modeling. By enabling robust length extrapolation through architectural innovation rather than extensive training, SWAN-GPT provides both immediate practical benefits for deployed models and a promising direction for future research into efficient context extension.

\bibliography{main}
\bibliographystyle{abbrv}

\appendix
\newpage
\section{Ablations}
\label{sec:ablations}
\begin{table*}[ht]
    \centering
    \begin{tabular}{l*{7}{S[table-format=1.3]}}
    \toprule
    Model & {512} & {1k} & {2k} & {4k} & {8k} & {16k} & {32k} \\
    \midrule
    local only & 1.000 & 0.601 & 0.285 & 0.127 & 0.057 & 0.022 & 0.010 \\
    global only (RoPE) & 1.000 & 0.985 & 0.000 & 0.000 & 0.000 & 0.000 & 0.000 \\
    global only (NoPE) & 1.000 & 1.000 & 0.000 & 0.000 & 0.000 & 0.000 & 0.000 \\
    global\_start (no scale) & 1.000 & 1.000 & 0.983 & 0.820 & 0.171 & 0.005 & 0.003 \\
    \midrule
    \midrule
    global\_start & 1.000 & 1.000 & 0.999 & 0.998 & 0.957 & 0.907 & 0.702 \\
    local\_start & 1.000 & 1.000 & 0.999 & 0.895 & 0.808 & 0.725 & 0.530 \\
    all\_global\_first & 1.000 & 0.599 & 0.316 & 0.113 & 0.044 & 0.017 & 0.010 \\
    all\_local\_first & 1.000 & 1.000 & 0.993 & 0.564 & 0.183 & 0.057 & 0.027 \\
    \bottomrule
    \end{tabular}
    \caption{NIAH scores across different context lengths for various SWAN configurations.}
    \label{tab:model-comparison}
\end{table*}

To investigate the impact of different hybrid attention configurations on length extrapolation capabilities, we conducted an ablation study using models with 0.5B parameters. Each model consisted of $24$ transformer decoder layers, with $16$ attention heads per layer, $1024$ hidden units, and a feedforward dimension of $4096$. We trained these models on a 350B token dataset using the AdamW optimizer, with a global batch size of $4096$. We employed a cosine decay learning rate schedule that peaked at $3e^{-3}$ after $2000$ warmup steps.  All sliding window attention layers used a window size of $512$ tokens with RoPE. For hybrid attention models we maintained a consistent $3$:$1$ ratio between local (sliding window) and global attention layers and used attention scaling during inference (though we include a control without attention scaling). Below is a brief description of each of the models:

\textbf{local only} - Implements sliding window attention across all layers.

\textbf{global only (RoPE)} - Standard transformer language model utilizing global attention with RoPE across all layers.

\textbf{global only (NoPE)} - Implements global attention with NoPE across all layers.

\textbf{global\_start} - Begins with a global NoPE layer followed by three consecutive sliding window layers, repeating this pattern throughout. For inference, we additionally evaluate a version without attention scaling to establish a baseline.

\textbf{local\_start} - Begins with three sliding window layers followed by a global NoPE layer, repeating this pattern throughout.

\textbf{all\_global\_first} - Concentrates all six global NoPE layers in the first positions, followed by sliding window layers.

\textbf{all\_local\_first} - Places all sliding window layers first, followed by six global NoPE layers.

\autoref{tab:model-comparison} shows results for the NIAH task from the RULER benchmark \cite{hsieh2024ruler}.\footnote{For simplicity we only evaluate the single NIAH task.} Among the baseline non-hybrid attention models, the \textbf{local only} model struggles to maintain high NIAH scores beyond its local window size ($512$), despite being trained on sequences of length $1$k. However, unlike the \textbf{global only} attention baselines (RoPE and NoPE), which completely fail beyond the training distribution, the \textbf{local only} model demonstrates a modest capacity for length extrapolation. In contrast, all hybrid attention variants show substantial improvements in generalizing beyond the training length.

When comparing the hybrid variants we find that interspersing global and local attention layers yields superior performance compared to grouping them together, as evidenced by the relatively poor performance of both \textbf{all\_global\_first} and \textbf{all\_local\_first} configurations. 
In particular, our best-performing model (\textbf{global\_start} achieves exceptional NIAH scores ($> 0.9$) at context lengths of $16$k --- 16 times the context length seen during training. It can also maintain robust performance (NIAH score $>0.7$) even at $32$k tokens, representing a 32-fold length extrapolation.

The critical role of attention scaling is demonstrated by our control experiment with \textbf{global\_start (no scale)}. While both scaled and unscaled variants maintain strong performance up to $2$k tokens, their behaviors diverge dramatically at longer contexts. The unscaled version shows rapid performance degradation beyond $4$k tokens, dropping from $0.820$ to $0.171$ at $8$k tokens and essentially failing ($0.005$) at $16$k tokens. In contrast, the scaled version maintains exceptional performance at $8$k tokens ($0.957$) and continues to achieve strong results at $16$k tokens ($0.907$), and even maintains moderately good results at $32$k tokens. This stark difference in length generalization --- $4$-fold extrapolation without scaling versus $32$-fold with scaling --- establishes attention scaling as a crucial mechanism for effective inference beyond the training length distribution. The graceful performance decline of the scaled model, compared to the abrupt deterioration of its unscaled counterpart, suggests that attention scaling helps maintain the model's ability to capture long-range dependencies even at extreme sequence lengths.

\section{Architecture \& Training}
Both RoPE-GPT-1B and SWAN-GPT-1B are trained from scratch with a batch size of 6M tokens (at 8k sequence length) with peak LR of 3e-3 for 1T tokens. We performed CPT for SWAN-8B with 32k sequence length and 6M token batch size at constant LR of 1e-5 for 300B tokens and ramped down to a LR of 5e-8 over another 15B tokens. Post-training for SWAN-8B model was performed in two stages. The first stage focused on a math and code blend with constant LR of 5e-6 followed by a second stage of general SFT at a constant LR of 1e-6.

\begin{table*}[ht]
\centering
\renewcommand{\arraystretch}{1.3}
\begin{tabular}{lcc}
\hline
\textbf{} & \textbf{SWAN-1B} & \textbf{SWAN-8B} \\
\hline
$n_{\text{layers}}$ & 24 & 32 \\
$d_{\text{model}}$ & 1536 & 4096 \\
$n_{\text{heads}}$ & 16 & 32 \\
$d_{\text{head}}$ & 96 & 128 \\
RoPE base & 1,000,000  & 1,000,000 \\
Normalization & RMSNorm & RMSNorm \\
global:local & 1:3 & 1:3 \\
SWA size & 512 & 512 \\

\hline
\end{tabular}
\caption{Architecture details for SWAN-1B and SWAN-8B models.}
\end{table*}

\end{document}